\documentclass[sigconf]{acmart}
\AtBeginDocument{%
  }
\usepackage{booktabs}   
\usepackage{multirow}   
\usepackage{multicol}   
\usepackage{caption}    
\usepackage{array}      
\usepackage{adjustbox}  
\usepackage{hyperref}
\usepackage{url}
\usepackage{pifont}
\usepackage{enumitem}
\usepackage{subcaption}
\usepackage{listings}
\usepackage{balance}

\usepackage{amsmath,amssymb,amsfonts,mathtools}
\usepackage{amsthm}
\usepackage[most]{tcolorbox}
\tcbset{colback=gray!3,colframe=gray!60}
\usepackage{amsmath,amssymb,amsthm}
\newtheorem{assumption}{Assumption}

\newtheorem{definition}{Definition}

\usepackage{mdframed}

\newmdenv[
  linewidth=0.6pt,
  skipabove=6pt,
  skipbelow=6pt,
  innertopmargin=6pt,
  innerbottommargin=6pt,
  innerleftmargin=6pt,
  innerrightmargin=6pt
]{promptframe}

\newcommand{\promptbox}[2]{%
\par\noindent\textbf{#1}\par
\begin{quote}
#2
\end{quote}
}

\copyrightyear{2026}
\acmYear{2026}
\setcopyright{cc}
\setcctype{by}
\acmConference[WWW '26]{Proceedings of the ACM Web Conference 2026}{April 13--17, 2026}{Dubai, United Arab Emirates}
\acmBooktitle{Proceedings of the ACM Web Conference 2026 (WWW '26), April 13--17, 2026, Dubai, United Arab Emirates}
\acmPrice{}
\acmDOI{10.1145/3774904.3792358}
\acmISBN{979-8-4007-2307-0/2026/04}

\begin{document}

\title[When to Trust: A Causality-Aware Calibration Framework for Accurate KG-RAG]{When to Trust: A Causality-Aware Calibration Framework for Accurate Knowledge Graph Retrieval-Augmented Generation}



\author{Jing Ren}
\authornote{Both authors contributed equally to this research.}
\orcid{0000-0003-0169-1491}
\affiliation{%
 \institution{RMIT University}
 \city{Melbourne}
 \country{Australia}
}
\email{jing.ren@ieee.org}

\author{Bowen Li}
\authornotemark[1]
\orcid{0009-0007-6470-5607}
\affiliation{%
 \institution{RMIT University}
 \city{Melbourne}
 \country{Australia}
}
\email{s3890442@student.rmit.edu.au}

\author{Ziqi Xu}
\authornote{Corresponding author.}
\orcid{0000-0003-1748-5801}
\affiliation{%
 \institution{RMIT University}
 \city{Melbourne}
 \country{Australia}
}
\email{ziqi.xu@rmit.edu.au}

\author{Xikun Zhang}
\orcid{0000-0002-0694-3654}
\affiliation{%
  \institution{RMIT University}
  \city{Melbourne}
  \country{Australia}
  }
\email{xikun.zhang@rmit.edu.au}

\author{Haytham Fayek}
\orcid{0000-0002-1840-7605}
\affiliation{%
  \institution{RMIT University}
  \city{Melbourne}
  \country{Australia}}
\email{haytham.fayek@ieee.org}

\author{Xiaodong Li}
\orcid{0000-0003-0346-1526}
\affiliation{%
  \institution{RMIT University}
  \city{Melbourne}
  \country{Australia}}
\email{xiaodong.li@rmit.edu.au}

\renewcommand{\shortauthors}{Jing Ren et al.}

\begin{abstract}
Knowledge Graph Retrieval-Augmented Generation (KG-RAG) extends the RAG paradigm by incorporating structured knowledge from knowledge graphs, enabling Large Language Models (LLMs) to perform more precise and explainable reasoning. While KG-RAG improves factual accuracy in complex tasks, existing KG-RAG models are often severely overconfident, producing high-confidence predictions even when retrieved sub-graphs are incomplete or unreliable, which raises concerns for deployment in high-stakes domains. To address this issue, we propose {Ca2KG}, a \underline{Ca}usality-aware \underline{Ca}libration framework for KG-RAG. Ca2KG integrates counterfactual prompting, which exposes retrieval-dependent uncertainties in knowledge quality and reasoning reliability, with a panel-based re-scoring mechanism that stabilises predictions across interventions. Extensive experiments on two complex QA datasets demonstrate that Ca2KG consistently improves calibration while maintaining or even enhancing predictive accuracy. 
The source code can be found at~\url{https://aisuko.github.io/ca2kg/}.
\end{abstract}




\maketitle

\section{Introduction}

\begin{figure}[t]
    \centering
    \begin{subfigure}{0.49\linewidth}
    \centering
    \includegraphics[width=\linewidth]{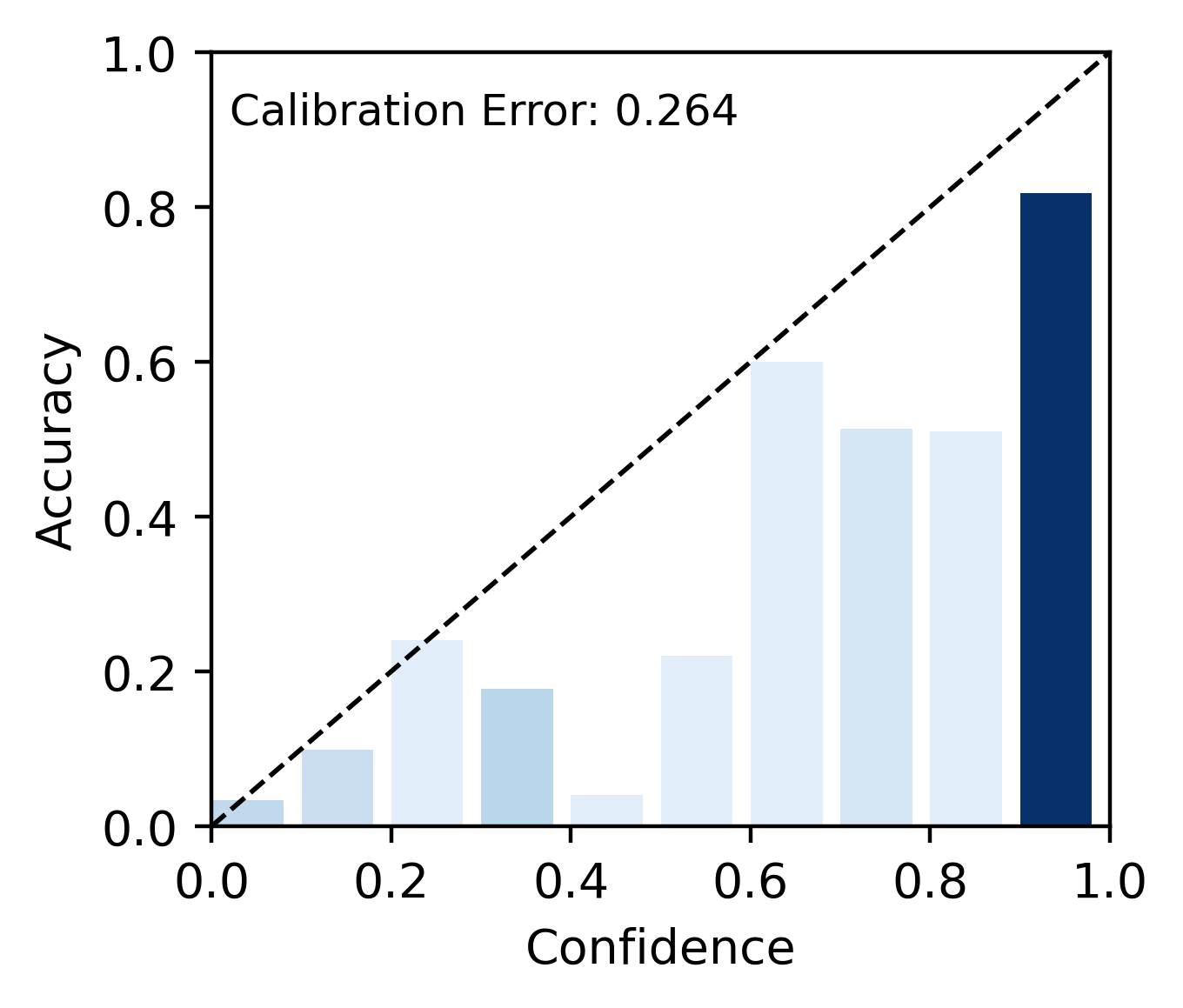}
    \label{fig:subfigA}
    \subcaption[]{na\"ive KG-RAG}
    \end{subfigure} 
    \hfill
    \begin{subfigure}{0.49\linewidth}
    \centering
    \includegraphics[width=\linewidth]{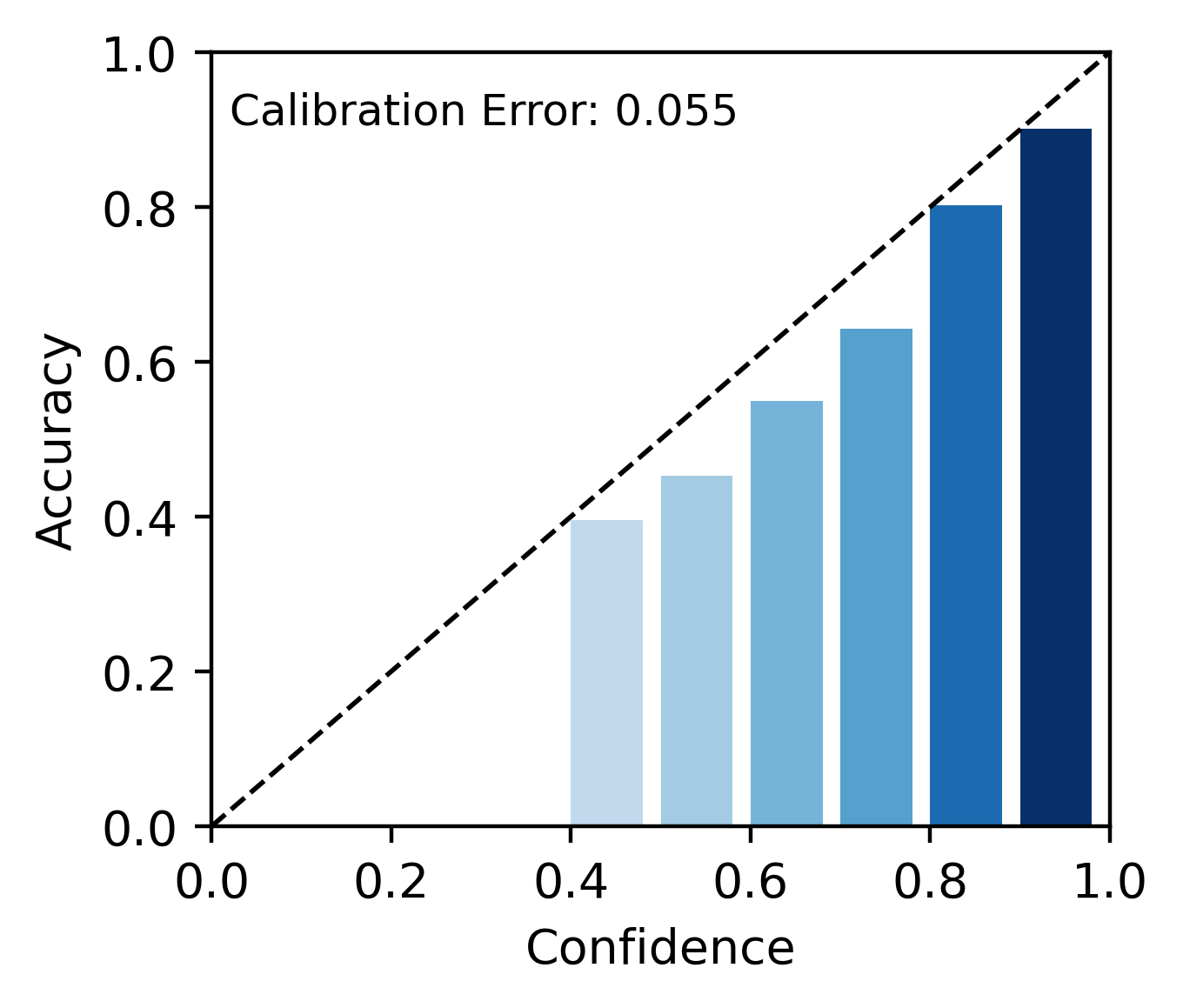}
    \label{fig:subfigB}
     \subcaption[]{Ca2KG (ours)}
    \end{subfigure}
    \caption{Calibration comparison on the MetaQA dataset. Calibration error quantifies the average discrepancy between a model's predicted confidence and its actual correctness. The na\"ive KG-RAG framework (a) is consistently over-confident with a high calibration error, whereas our Ca2KG framework (b) achieves much better calibration with a reduced error. 
    }
    \label{fig:fig1}
\end{figure}

Retrieval-Augmented Generation (RAG) is a powerful framework that enhances Large Language Models (LLMs) by retrieving relevant external information from large-scale web corpora to support more informed and factual generation~\cite{gao2023retrieval,zhu2023large,ram2023context,izacard2023atlas}. By integrating retrieval and generation into a unified pipeline, RAG enables LLMs to access knowledge beyond their training data and mitigates hallucination by grounding responses in retrieved web evidence~\cite{lewis2020retrieval,salemi2024evaluating,yu2024rankrag}. However, conventional RAG systems rely primarily on unstructured web text, which often lacks the semantic precision, logical structure, and interpretability necessary for complex reasoning tasks. To overcome these limitations, Knowledge Graph Retrieval-Augmented Generation (KG-RAG) extends the RAG paradigm by incorporating structured knowledge from web-based knowledge graphs. This integration allows models to retrieve and reason over multi-hop semantic paths, leading to more precise and explainable generation, especially for web-scale tasks requiring multi-step reasoning~\cite{ren2023graph,zhang2025survey,xiang2025use,liang2025kag}.


Although KG-RAG has been shown to substantially improve factual accuracy in complex tasks~\cite{xiang2025use,kim2023kg,liu2024knowledge,xia2026graph}, the na\"ive KG-RAG framework often exhibits such over-confidence despite factual inaccuracies (see Figure~\ref{fig:fig1}). This issue directly reflects poor calibration, which measures how well a model’s predicted confidence aligns with its actual likelihood of being correct. A well-calibrated model outputs high confidence only when its predictions are reliable, and low confidence when uncertainty is warranted. Calibration is especially critical for KG-RAG because inaccurate confidence can mislead downstream reasoning steps, amplify retrieval errors, and cause incorrect entity linking or graph traversal. Therefore, improving calibration is urgently needed for reliable web-scale knowledge retrieval and generation.

Building on this need, prior research has investigated calibration as a core aspect of RAG system reliability, along with broader uncertainty estimation, which seeks to quantify and decompose different sources of predictive uncertainty~\cite{chang2024survey,yang2024alignment,steyvers2025large}. However, these efforts have primarily focused on standard LLMs or text-based RAG models, with limited exploration of KG-RAG. Unlike conventional RAG that retrieves unstructured web text, KG-RAG relies on structured knowledge from web-based knowledge graphs in the form of entities, relations, and multi-hop paths, which introduces unique challenges in both retrieval and reasoning. These characteristics raise a central research question: \textit{how can we improve calibration in KG-RAG by explicitly modelling retrieval-dependent uncertainties, so that the model’s confidence better reflects its true reliability?}

To address this problem, we propose {Ca2KG}, a \underline{Ca}usality-aware \underline{Ca}libration framework for \underline{K}nowledge Graph Retrieval-Augmented \underline{G}eneration that combines counterfactual prompting with panel-based re-scoring. The framework is motivated by the causal view that different prompting interventions can be regarded as treatments that expose retrieval-dependent uncertainties. Inspired by the counterfactual prompting framework~\cite{chen2024controlling}, we design prompts that simulate alternative web retrieval scenarios, such as “suppose the wrong knowledge path was selected” or “suppose the reasoning over the retrieved path was flawed”, thereby encouraging the framework to introspect on the robustness of retrieved evidence. Building on these counterfactual generations, we introduce a panel-based re-scoring process, where the framework evaluates candidate answers under each intervention and assigns calibrated probabilities through a unified scoring scheme. 
This two-stage process allows us to not only expose and quantify uncertainties arising from retrieved sub-graphs, but also to stabilise predictions across interventions, ultimately leading to more calibrated and reliable web-scale KG-RAG systems. 
In summary, our contributions are as follows:
\begin{itemize}[leftmargin=0.5cm]
    \item We present the first systematic study on the calibration of KG-RAG systems, revealing that existing frameworks often produce severely overconfident predictions. 
    \item We propose {Ca2KG}, a causality-aware calibration framework that integrates counterfactual prompting with panel-based re-scoring, and introduce a stability-based scoring mechanism that explicitly accounts for retrieval-dependent uncertainties in both knowledge quality and reasoning reliability. 
    \item We conduct extensive experiments on two complex QA datasets, demonstrating that Ca2KG consistently improves calibration metrics (e.g., Expected Calibration Error, Brier Score) while maintaining or even enhancing predictive accuracy. 
\end{itemize}

Our work falls under the \textit{Semantics and Knowledge} track as it focuses on calibrating KG-RAG, which directly relies on Web-based structured knowledge graphs with machine-interpretable semantics. Our contribution advances frameworks that synergise knowledge graphs and LLMs, leading to more trustworthy semantic reasoning and user-facing applications on the Web.

\begin{figure*}[t]
    \centering
    \includegraphics[width=0.99\textwidth]{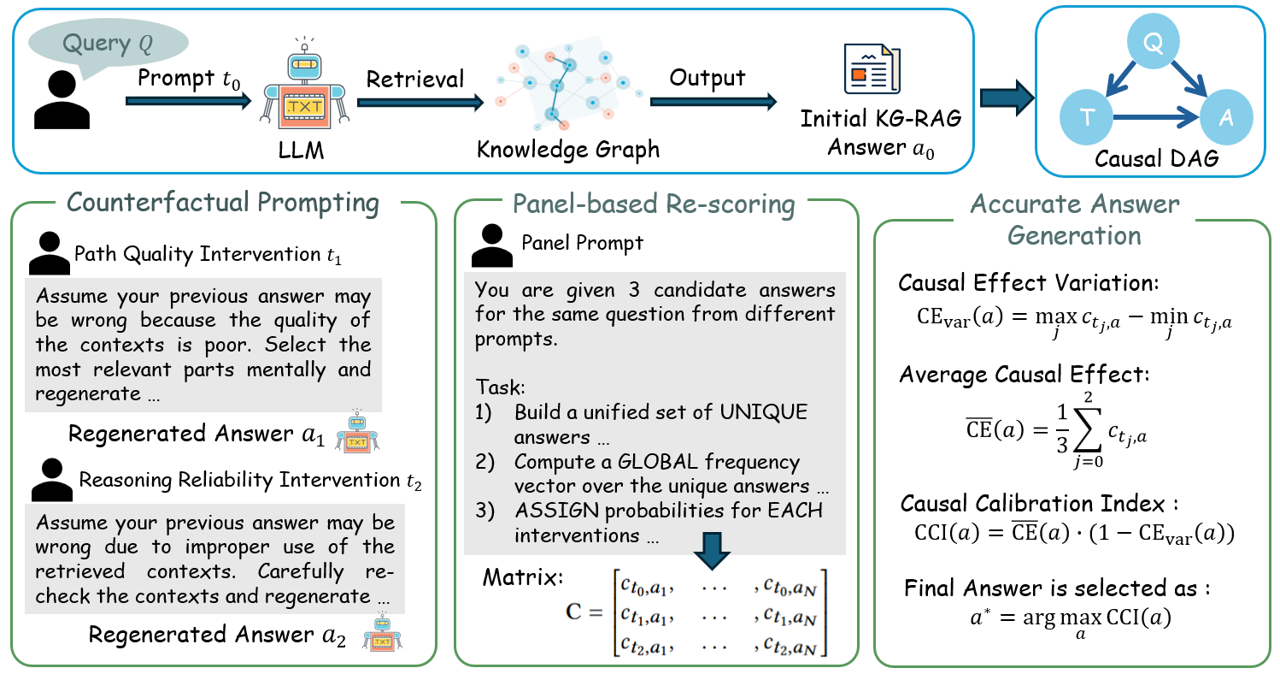}
    \caption{The overall architecture of Ca2KG. Given a query, the initial KG-RAG pipeline produces a baseline answer. Counterfactual prompting introduces interventions on retrieved paths, simulating quality and reasoning failures to generate alternative answers. Panel-based re-scoring evaluates all candidates under each prompt, forming a $3 \times N$ probability matrix. Finally, the Causal Calibration Index (CCI) combines support and stability across interventions to select the final calibrated answer.}
    \label{fig:framework}
\end{figure*}

\section{Preliminaries}

\subsection{Structural Causal Model}
\label{subsec:scm}

A Structural Causal Model (SCM) formalises causal relationships between variables using a directed acyclic graph (DAG). Formally, an SCM is represented by a causal DAG $\mathcal{G}=(\mathcal{V},\mathcal{E})$, where each node $X \in \mathcal{V}$ corresponds to a random variable, and each directed edge $(U \!\to\! V)\in\mathcal{E}$ encodes a direct causal influence of $U$ on $V$. For a node $X$, we denote its set of parents by $\mathrm{PA}(X)$, which are all nodes with edges pointing into $X$. 

A causal path from $X$ to $Y$ is a directed sequence of nodes $X \to \cdots \to Y$ that follows the arrow direction, indicating that $X$ is a (possibly indirect) cause of $Y$. This graphical representation provides the foundation for reasoning about intervention and counterfactual in causal inference.

\subsection{Causal Intervention}
\label{subsec:do}

In real-world applications, it is essential to distinguish between mere statistical associations and genuine causal relationships. The observational conditional distribution $P(Y \mid X)$ characterises how two variables co-vary in passively collected data, but it does not reveal what would happen to $Y$ if $X$ was actively manipulated. To address this limitation, the \emph{do-operator}~\citep{pearl2009causality} is introduced, formalising the notion of a causal intervention.

The operator $\mathrm{do}(X=x)$ represents a surgical intervention that forces the variable $X$ to take the value $x$ while cutting all incoming edges into $X$ in the causal DAG $\mathcal{G}$. Intuitively, this breaks the natural causal mechanisms that determine $X$ and replaces them with a fixed assignment. As a result, variation in $X$ no longer depends on its original causes, but solely on the external manipulation. 

Formally, given a causal DAG $\mathcal{G}=(\mathcal{V},\mathcal{E})$ and its Markov factorisation.
\begin{equation}
    P(\mathcal{V}) = \prod_{V\in\mathcal{V}} P\bigl(V \mid \mathrm{PA}(V)\bigr),
\end{equation}
the distribution after an intervention $\mathrm{do}(X=x)$ can be expressed as follows:
\begin{equation}
P(\mathcal{V}\setminus\{X\} \mid \mathrm{do}(X=x))
= \prod_{V\in \mathcal{V}\setminus\{X\}} P\bigl(V \mid \mathrm{Pa}(V)\bigr)\Big|_{X=x}.
\end{equation}

Based on this formula, we can formalise the \emph{causal effect} of a treatment variable $T$ on an outcome variable $Y$ as the following mapping:
\begin{equation}
t \mapsto P(Y \mid \mathrm{do}(T=t)),
\end{equation} which specifies the distribution of $Y$ that would arise under different interventions on $T$. 

A common summary measure is the difference in expectations under two interventions, 
\begin{equation}
\mathbb{E}[Y \mid \mathrm{do}(T=t')] - \mathbb{E}[Y \mid \mathrm{do}(T=t'')],
\end{equation}
which quantifies how outcomes would change when the treatment is shifted from $t''$ to $t'$.

However, causal effects cannot in general be computed directly from observational data, since multiple causal models may give rise to the same joint distribution. This motivates the notion of \emph{identifiability}, which requires that a causal effect $P(Y \mid \mathrm{do}(T=t))$ be uniquely determined from the observational distribution together with the causal graph. Identifiability ensures that the effect of interventions can be inferred from observed data under appropriate assumptions about the underlying causal structure.

These notions of interventions and causal effects provide the foundation of our framework. Specifically, we interpret different counterfactual prompting strategies as distinct treatments, and regard the resulting changes in model outputs as causal effects. This causal perspective enables us to move beyond surface-level associations in observed responses and to systematically quantify how interventions on prompts influence model predictions.

\section{Methodology}
In this section, we present {Ca2KG}, our causality-aware calibration framework designed to improve the reliability of KG-RAG. The overall architecture is illustrated in Figure~\ref{fig:framework}. Our framework is structured around four main components: (i) formulating the calibration task within a causal inference perspective, (ii) designing counterfactual prompting strategies to simulate retrieval and reasoning failures, (iii) employing a panel-based re-scoring mechanism to estimate interventional distributions, and (iv) developing a causal calibration criterion for final answer selection. Together, these components enable the model to generate predictions that are both accurate and well-calibrated.

\subsection{Problem Statement}
We consider the calibration task of KG-RAG, where an LLM is augmented with structured knowledge from a KG to generate accurate and grounded responses to queries. Formally, given a query $q$, the system first performs entity linking to identify the relevant KG entities $\mathcal{E}_q = \{e_1, \ldots, e_m\}$. Based on these entities, a sub-graph $\mathcal{G}^q_{KG} = (V_q, E_q)$ is retrieved, which contains candidate paths $\mathcal{P}_q = \{p_1, \ldots, p_k\}$ that may support answering $q$. Each path $p_i$ can be represented as a sequence of entity-relation-entity triples or as natural language statements, and is encoded into a prompt $t_{\text{prompt}}$. The final input to the LLM is constructed as $ [q, t_{\text{prompt}}]$, from which the model $f(\cdot)$ generates an answer $a$ together with a confidence score $c \in [0,1]$. However, such confidence estimates are often unreliable, particularly when the retrieved knowledge is incomplete, noisy, or biased. This motivates us to propose a causal prompting framework to enhance the reliability of confidence scores in KG-RAG by incorporating causal-aware reasoning, thereby enabling more trustworthy answer selection.

\subsection{Causal Principles}
We conceptually formalise the relationship between the query $Q$, the prompt $T$, and the answer $A$ using a causal DAG $\mathcal{G}$. As illustrated in Figure~\ref{fig:framework}, $\mathcal{G}$ contains three directed edges:
\begin{equation}
    T \to A, \quad Q \to T, \quad Q \to A.
\end{equation}

Here, $Q$ denotes the query, $T$ represents the prompt intervention, and $A$ denotes the candidate answer. This structure reflects that the query influences both the chosen prompt and the resulting answer, while the prompt itself also has a direct causal effect on the answer. 

In our framework, we focus on heterogeneous causal DAGs. For a given query $q$, we design counterfactual prompting strategies $t_n$ that serve as treatments, and our goal is to obtain unbiased estimates of the causal effects of these treatments on the resulting answers. These causal effects can further be interpreted as calibrated confidence scores, which provide a principled basis for reliable answer selection and subsequent generation.

Within this causal structure, the path $T \leftarrow Q \to A$ serves as a confounding path. As a result, the observational distribution $P(A \mid T)$ cannot be directly interpreted as the causal effect of $T$ on $A$, since both are influenced by the query $Q$. Without adjusting for $Q$, any estimated effect of prompts on answers would be biased. To achieve causal identification, we adopt three standard assumptions from causal inference for KG-RAG:

\begin{assumption}[Stable Unit Treatment Value Assumption (SUTVA)]
\label{ass:SUTVA}
    For KG-RAG, the potential answers associated with a given query are unaffected by the prompting strategies applied to other queries (no interference). Moreover, each prompting strategy is assumed to have a unique, well-defined version for that query, such that different forms of the same strategy do not lead to different potential answers (consistency).
\end{assumption}

This assumption has two key principles. First, \emph{no interference}: the potential answer for one query is unaffected by prompting strategies applied to other queries, provided queries are evaluated independently. Second, \emph{consistency}: each prompting strategy must correspond to a single, well-defined generative process for a given query, without ambiguity or hidden variations. In practice, since LLMs often operate with stochastic decoding (e.g., non-zero temperature), consistency is interpreted at the level of the induced output distribution rather than individual sampled responses.

\begin{assumption}[Unconfoundedness]
\label{ass:Unconfoundedness}
    For KG-RAG, the potential answers are assumed to be conditionally independent of the prompting strategy $T$ given the query $Q$. 
    Formally,
    \begin{equation}
    A \perp\!\!\!\perp T \mid Q. 
    \end{equation}
\end{assumption}

This assumption implies that, once the query is taken into account, the assignment of prompting strategies does not carry any additional hidden confounding information about the answers.

\begin{assumption}[Overlap]
    \label{ass:Overlap}
    For KG-RAG, any query $Q=q$, every prompting strategy $t$ has a non-zero probability of being assigned. Formally,
    \begin{equation}
    0 < P(T = t \mid Q = q) < 1. 
    \end{equation}
\end{assumption}

This assumption ensures that all prompting strategies are feasible for each query, making causal comparisons across strategies possible.

Together, these assumptions guarantee that causal effects are, in principle, identifiable from observational data. To operationalise this identification in our framework, we now introduce {back-door criterion}~\citep{pearl2009causality}, which provides a graphical condition for determining suitable adjustment sets in a causal DAG.

\begin{definition}[Back-Door Criterion]
\label{def:BD}
A set of variables ${Z}$ satisfies the back-door criterion relative to an ordered pair of variables $(T,A)$ in a causal DAG $\mathcal{G}$ if: 
\begin{enumerate}
    \item no node in ${Z}$ is a descendant of $T$; and 
    \item ${Z}$ blocks every path between $T$ and $A$ that contains an arrow into $T$.
\end{enumerate}
\end{definition}

By adjusting for such a set ${Z}$, the causal effect of $T$ on $A$ can be identified as
\begin{equation}
P(Y \mid do(T)) = \sum_{z} P(A \mid T, Z=z)\, P(Z=z).
\end{equation}

In our causal DAG, the query $Q$ satisfies the back-door criterion relative to the treatment–outcome pair $(T, A)$, where $T$ denotes the prompt and $A$ the answer. It is not a descendant of $T$, and it blocks the only back-door path $T \leftarrow Q \to A$, which implies that $A \perp\!\!\!\perp T \mid Q$. Therefore, by conditioning on $Q$, the causal effect of prompt strategies $T$ on the answer $A$ can be identified as
\begin{equation}
P(A \mid do(T)) = P(A \mid T, Q).
\end{equation}

Note that we focus on heterogeneous causal DAGs where each query $q$ corresponds to a distinct causal structure. Thus, we are interested in query-specific causal effects rather than population-averaged effects, and no marginalisation over $Q$ is required.

\subsection{Counterfactual Prompting}
We aim to design a counterfactual prompting strategy that perturbs KG-related evidence in two ways: by varying the quality of retrieved paths and by altering their usage strategy. This strategy encourages the model to reflect on its confidence under such interventions. The core challenge is to assess model uncertainty without relying on gold-standard answers, which is crucial for enabling selective prediction and risk-aware decision-making in downstream applications. Motivated by cognitive theories of counterfactual reasoning~\citep{pearl2009causality}, we propose two counterfactual prompting strategies that systematically simulate failure scenarios in KG-RAG systems.

\begin{itemize}[leftmargin=0.5cm]
    \item \textbf{Path Quality Intervention ($t_{1}$)}: simulates scenarios where the retrieved KG paths are irrelevant, incomplete, or noisy, and prompts the LLM accordingly: ``Assume your previous answer is wrong because the quality of the referred contexts is poor. Re-select the most relevant parts from the given contexts and regenerate the answer using one or a few words. Output MUST be exactly one line in this format: \{final answer\}. Do not include any other text. Examples: \{Italian Languages\}''
    \item \textbf{Reasoning Reliability Intervention ($t_{2}$)}: simulates scenarios where the model’s reasoning over otherwise valid KG paths may be unreliable or flawed, and prompts the LLM accordingly: ``Assume your previous answer is wrong due to improper use of the retrieved contexts. Carefully re-check the provided contexts and regenerate the answer using one or a few words. Output MUST be exactly one line in this format: \{final answer\}. Do not include any other text. Examples: \{Italian Languages\}''
\end{itemize}

Each counterfactual prompt $t_{1}$ and $t_{2}$ is fed into the same backbone LLM, which produces the corresponding counterfactual answers $a_{1}$ and $a_{2}$. In addition to the counterfactual prompting strategies, we also include an initial prompt $t_{0}$ as the baseline, which generates the initial answer $a_{0}$. The prompt is defined as follows: ``Use the provided contexts to answer the question. If the contexts are incomplete or weak, still provide your best possible answer. Output MUST be exactly one line in this format: \{final answer\}. Do not include any other text. Examples: \{Italian\}''

\subsection{Panel-based Re-scoring}
\label{sec:panel}

To estimate the interventional distribution $P(A \mid do(T))$, we introduce a unified Panel Prompt, which serves as an evaluator and prompts the framework as follows:

\promptbox{}{
You are given 3 candidate answers for the same query from different prompts: $t_{0}$: $a_0$; $t_{1}$: $a_1$; $t_{2}$: $a_2$. 

\textbf{Task:}

\begin{enumerate}[label=(\arabic*), leftmargin=0.6cm]
  \item Build a unified set of UNIQUE answers by merging semantically identical strings across $a_0$, $a_1$, $a_2$.
  \item Compute a GLOBAL frequency vector over the unique answers based on how many of $a_0$/$a_1$/$a_2$ map to each canonical answer.
  \item ASSIGN probabilities for EACH interventions ($t_{0}$, $t_{1}$, $t_{2}$) to be EXACTLY this global frequency vector normalised by 3 (counts/3), after duplicate aggregation and before rounding.
\end{enumerate}

\medskip
\textbf{Re-scoring Structure:}

{Merging Rules}
\begin{itemize}[leftmargin=0.5cm]
  \item Ignore case, whitespace, punctuation, trivial formatting, and plural/singular differences.
  \item Choose a clean canonical form for answers (e.g., title case).
  \item If multiple inputs ($a_0$/$a_1$/$a_2$) map to the same canonical answer, aggregate them by SUMMING that answer's frequency before normalisation.
\end{itemize}

{Probability rules (HARD CONSTRAINTS)}
\begin{itemize}[leftmargin=0.5cm]
  \item  Let count[i] be how many of {$a_0$/$a_1$/$a_2$} map to answers[i]. Then for every intervention T in {$t_{0}$, $t_{1}$, $t_{2}$}, set T[i] = round(count[i]/3, 2).
  \item Do NOT output equal probabilities across answers when counts differ.
  \item  Each probability $\in [0.00, 1.00]$; sums may be < 1.00 after rounding.
\item  If an intervention has no plausible answers, use zeros. Return EXACTLY one line of STRICT JSON, no wrapper, no extra text.
\end{itemize}
}

Given a specific query $q$, we consider the candidate answer set $A=\{a_0,\dots,a_N\}$ obtained by aggregating outputs from the three prompt strategies $T=\{t_0,t_1,t_2\}$. Since each intervention produces one answer, the candidate set size satisfies $0 \leq N \leq 2$, depending on whether duplicates are merged during aggregation. The Panel Prompt then re-scores each candidate answer $a_i \in A$ under every prompt strategy $t_j \in T$.

Formally, for each prompt strategy $t_j \in \{t_0,t_1,t_2\}$, the Panel Prompt produces a probability distribution over the candidate answer set $A$: 
\begin{equation}
    P(A=a_i \mid T=t_j, Q), \quad i=1,\dots,N,\; j=0,1,2.
\end{equation}

These probabilities can be interpreted as the causal effect of prompt strategy $t_j$ on candidate answer $a_i$ under query $Q$. Collecting these causal effects yields a score matrix $\mathbf{C} \in \mathbb{R}^{3 \times N}$, where each row corresponds to a prompt strategy $t_j$ and each column corresponds to a candidate answer $a_i$. The matrix is given as:
\begin{equation}
\mathbf{C} = 
\begin{bmatrix}
c_{t_0,a_1}, &\dots& ,c_{t_0,a_N} \\
c_{t_1,a_1}, &\dots& ,c_{t_1,a_N} \\
c_{t_2,a_1}, &\dots& ,c_{t_2,a_N}
\end{bmatrix}.
\end{equation}

\subsection{Accurate Answer Generation}

Each candidate answer $a \in A$ receives probability assignments, interpreted as causal effects, across prompt strategies. However, selecting the candidate with the highest average probability may yield unstable decisions. To address this, we design a stability-aware selection criterion that balances accuracy with robustness under interventions.

We first measure the variability of the probabilities assigned to each candidate across prompt strategies. Given the probability matrix $\mathbf{C}$, the causal effect variation ($\text{CE}_{\text{var}}$) of a candidate answer $a$ is defined as:
\begin{equation}
\text{CE}_{\text{var}}(a) = \max_{j} c_{t_j,a} - \min_{j} c_{t_j,a},\; j=0,1,2
\end{equation}
where $c_{t_j,a}$ denotes the probability of candidate $a$ under treatment $t_j$. A higher $\text{CE}_{\text{var}}(a)$ indicates greater instability across prompt strategies.

Next, we compute the average causal effect $\text{CE}$ for candidate $a$:
\begin{equation}
\overline{\text{CE}}(a) = \frac{1}{3} \sum_{j=0}^2 c_{t_j,a}.
\end{equation}

To jointly capture accuracy and calibration, we define the Causal Calibration Index (CCI):
\begin{equation}
\text{CCI}(a) = \overline{\text{CE}}(a) \cdot (1 - \text{CE}_{\text{var}}(a)),
\end{equation}
which promotes candidates with both high average probability and consistent behaviour across interventions, thereby improving answer selection and calibration.

The final answer is selected as:
\begin{equation}
a^* = \arg\max_a \text{CCI}(a).
\end{equation}

This design ensures that the selected answer is not only highly probable but also well-calibrated across interventions, leading to more reliable causal decision-making.

\section{Experimental Setup}
\subsection{Datasets}
We evaluate our method on two widely used benchmarks for multi-hop question answering over knowledge graphs, MetaQA~\cite{zhang2018variational} and WebQSP~\cite{yih2016value}, each supporting 1-hop and 3-hop reasoning tasks. This setup allows us to assess model performance under both shallow and deep reasoning settings. 

{MetaQA} is a synthetic, movie-domain dataset containing over 400K natural language questions generated from a structured KG with entities such as movies, actors, directors, and genres. {WebQSP} is a real-world dataset extending WebQuestions with semantic parses grounded to Freebase. It contains over 4,700 user queries annotated with SPARQL-like logical forms, supporting multi-hop reasoning and entity linking. Compared to MetaQA, WebQSP is more challenging due to its natural language variability, diverse entities and relations, and reliance on precise semantic parsing. 

\subsection{Baselines}
We select a series of baselines with different prompting strategies and calibration methods to compare against our proposed framework. Specifically, we consider three prompt-based baselines: 
If-or-Else (IoE) prompting framework~\cite{li2024confidence} improves self-correction by leveraging the IoE prompting principle to adjust responses based on model confidence.  
Self-Correct~\cite{huang2023large} explores the role and effectiveness of self-correction in LLMs through a three-step prompting strategy.  
RC-RAG~\cite{chen2024controlling} mitigates risks in LLMs by enforcing consistency in answers and discarding inconsistent ones.  


We also adopt three verbalised strategies for extracting confidence estimates, following~\cite{tian2023just}: 
Verb1S-top4 prompts the model to produce four candidate answers and assign a probability of correctness to each within a single response.  Verb2S-top4 separates the process into two stages: the first turn generates four candidate answers, and the second turn elicits correctness probabilities for each.  
Verb2S-CoT adds chain-of-thought reasoning in the first turn before producing a single answer, while the second turn requests a confidence score for that answer with the reasoning retained in context.  


\subsection{Backbone LLMs}
In our experiments, we employ two backbone LLMs with different parameter scales and accessibility: LLaMA-3~\cite{Grattafiori2024Llama3Herd} and GPT-3.5~\cite{openai_chatgpt_2022}. This selection enables evaluation of calibration performance across both open- and closed-source models, as well as across model sizes, providing a comprehensive and balanced experimental setting.

\subsection{Metrics} 
In our experiments, we use \textbf{Accuracy} (Acc) as the primary performance metric. Following~\cite{tian2023just}, we further assess model calibration with a range of established metrics.
Acc measures the proportion of correctly predicted labels over the total number of test instances and directly reflects the discriminative ability of the model.
\textbf{Expected Calibration Error} (ECE) is a widely used metric for quantifying the alignment between predicted confidence and empirical accuracy~\cite{guo2017calibration}. 
To compute ECE, predictions are first partitioned into $M$ equally spaced bins based on their confidence scores. For each bin $B_m$, the average confidence $\text{conf}(B_m)$ and the empirical accuracy $\text{acc}(B_m)$ are computed. 
ECE is defined as the weighted average of the absolute differences between confidence and accuracy across all bins: $\text{ECE} = \sum_{m=1}^{M} \frac{|B_m|}{n} \left| \text{acc}(B_m) - \text{conf}(B_m) \right|$, where $|B_m|$ is the number of samples in bin $m$ and $n$ is the total number of samples. A lower ECE indicates better calibration, meaning predicted probabilities more accurately reflect the true likelihood of correctness.
\textbf{Brier Score} (BS) provides a complementary perspective on calibration by measuring the mean squared difference between predicted confidence and the ground-truth label. 
$\text{BS} = \frac{1}{n} \sum_{i=1}^{n} (c_i - y_i)^2$,
where $c_i \in [0, 1]$ is the predicted confidence score and $y_i \in \{0, 1\}$ is the ground-truth indicator of correctness. A lower BS indicates that predicted probabilities are closer to the true outcomes.
In addition to standard calibration metrics, we also report the \textbf{Area Under the Selective Accuracy-Coverage Curve} (AUC), introduced in~\cite{geifman2017selective}. This metric evaluates the trade-off between accuracy and coverage when the model abstains from predictions with low confidence. By integrating accuracy across different coverage levels, AUC captures the ability of the model to identify and withhold uncertain predictions, offering a complementary perspective on reliability beyond ECE and BS.

\begin{table*}[t]
\centering
\caption{Results on MetaQA and WebQSP. KG-RAG denotes the standard retrieval-augmented generation baseline without any calibration adjustment. The symbol ``--'' indicates that calibration metrics are not reported, since baselines such as IoE, Self-Correct, and RC-RAG are not originally designed to produce probabilistic confidence estimates. Best results are highlighted in bold.}
\setlength{\tabcolsep}{3pt}
\begin{tabular}{l cccc cccc cccc cccc} 
\toprule
 & \multicolumn{8}{c}{\textbf{MetaQA}} & \multicolumn{8}{c}{\textbf{WebQSP}} \\
\cmidrule(lr){2-9}\cmidrule(lr){10-17}
 & \multicolumn{4}{c}{\textbf{1-hop}} & \multicolumn{4}{c}{\textbf{3-hop}} 
 & \multicolumn{4}{c}{\textbf{1-hop}} & \multicolumn{4}{c}{\textbf{3-hop}} \\
\cmidrule(lr){2-5}\cmidrule(lr){6-9}\cmidrule(lr){10-13}\cmidrule(lr){14-17}

\textbf{Method} & Acc$\uparrow$ & ECE$\downarrow$ & BS$\downarrow$ & AUC$\uparrow$ 
                & Acc$\uparrow$ & ECE$\downarrow$ & BS$\downarrow$ & AUC$\uparrow$ 
                & Acc$\uparrow$ & ECE$\downarrow$ & BS$\downarrow$ & AUC$\uparrow$ 
                & Acc$\uparrow$ & ECE$\downarrow$ & BS$\downarrow$ & AUC$\uparrow$ \\
\midrule
\multicolumn{17}{c}{\textbf{GPT-3.5}} \\
\midrule
KG-RAG           & 0.554 & 0.433 & 0.416 & 0.706 & 0.177 & 0.821 & 0.819 & 0.116 
                  & 0.340 & 0.646 & 0.538 & 0.529 & 0.031 & 0.908 & 0.911 & 0.050 \\
IoE               & 0.843 &   -   &   -   &   -   & 0.590 &   -   &   -   &   -   
                  & 0.557 &   -   &   -   &   -   & 0.673 &   -   &   -   &   -   \\
Self-Correct      & 0.539 &   -   &   -   &   -   & 0.219 &   -   &   -   &   -   
                  & 0.409 &   -   &   -   &   -   & 0.162 &   -   &   -   &   -   \\
RC-RAG            & 0.874 & 0.112 & 0.094 & 0.934 & 0.815 & 0.084 & 0.083 & 0.943 
                  & 0.704 & 0.302 & 0.295 & 0.727 & 0.794 & 0.139 & 0.144 & 0.863 \\
Verb1S-Top4       & 0.813 & 0.160 & 0.152 & 0.897 & 0.837 & 0.094 & 0.127 & 0.889 
                  & 0.618 & 0.309 & 0.313 & 0.724 & 0.492 & 0.449 & 0.448 & 0.437 \\
Verb2S-Top4       & 0.829 & 0.187 & 0.130 & 0.931 & 0.861 & 0.190 & 0.107 & 0.921 
                  & 0.458 & 0.291 & 0.285 & 0.731 & 0.176 & 0.676 & 0.586 & 0.155 \\
Verb2S-CoT        & 0.562 & 0.437 & 0.437 & 0.618 & 0.610 & 0.520 & 0.471 & 0.533 
                  & 0.548 & 0.420 & 0.401 & 0.593 & 0.670 & 0.539 & 0.434 & 0.486 \\
Ca2KG  & \textbf{0.876} & \textbf{0.067} & \textbf{0.078} & \textbf{0.947} 
                  & \textbf{0.896} & \textbf{0.055} & \textbf{0.069} & \textbf{0.950} 
                  & \textbf{0.769} & \textbf{0.196} & \textbf{0.184} & \textbf{0.837} 
                  & \textbf{0.819} & \textbf{0.108} & \textbf{0.128} & \textbf{0.920} \\
\midrule
\multicolumn{17}{c}{\textbf{LLaMA-3}} \\
\midrule
KG-RAG            & 0.724 & 0.264 & 0.220 & 0.887 & 0.470 & 0.242 & 0.269 & 0.697 
                  & 0.579 & 0.330 & 0.301 & 0.710 & 0.004 & 0.739 & 0.551 & 0.004 \\
IoE               & 0.808 &   -   &   -   &   -   & 0.710 &   -   &   -   &   -   
                  & 0.372 &   -   &   -   &   -   & 0.606 &   -   &   -   &   -   \\
Self-Correct      & 0.605 &   -   &   -   &   -   & 0.348 &   -   &   -   &   -   
                  & 0.455 &   -   &   -   &   -   & 0.165 &   -   &   -   &   -   \\
RC-RAG            & 0.812 & 0.124 & 0.118 & 0.912 & 0.736 & 0.118 & 0.121 & 0.913 
                  & 0.642 & 0.269 & 0.258 & 0.759 & 0.315 & 0.354 & 0.368 & 0.392 \\
Verb1S-Top4       & 0.646 & 0.108 & 0.126 & 0.933 & 0.595 & 0.091 & 0.086 & 0.941 
                  & 0.358 & 0.456 & 0.436 & 0.544 & 0.397 & 0.486 & 0.481 & 0.445 \\
Verb2S-Top4       & 0.834 & 0.147 & 0.118 & 0.921 & 0.736 & 0.118 & 0.121 & 0.913 
                  & 0.642 & 0.269 & 0.258 & 0.759 & 0.315 & 0.354 & 0.368 & 0.392 \\
Verb2S-CoT        & 0.668 & 0.373 & 0.282 & 0.844 & 0.648 & 0.320 & 0.288 & 0.753 
                  & 0.324 & 0.630 & 0.577 & 0.433 & 0.501 & 0.619 & 0.602 & 0.392 \\
Ca2KG         & \textbf{0.861} & \textbf{0.085} & \textbf{0.093} & \textbf{0.944} 
                  & \textbf{0.872} & \textbf{0.064} & \textbf{0.079} & \textbf{0.952} 
                  & \textbf{0.738} & \textbf{0.211} & \textbf{0.203} & \textbf{0.812} 
                  & \textbf{0.612} & \textbf{0.242} & \textbf{0.266} & \textbf{0.657} \\
\bottomrule
\end{tabular}
\label{tab:main2}
\end{table*}
 
\section{Experimental Results}
We structure our investigation around the following five research questions: RQ1: How does {Ca2KG} improve calibration relative to baselines, and how does this translate into downstream utility (e.g., accuracy)? RQ2: How does the capability of the backbone LLM affect the overall performance? RQ3: How does reasoning complexity (e.g., 1-hop vs. 3-hop) influence both calibration and accuracy? RQ4: What is the computational cost of the proposed {Ca2KG} framework? RQ5: What are the respective contributions of the two counterfactual prompting strategies to model effectiveness?  

In addition, for each dataset used in our experiments, we provide a case study in Appendix~\ref{Case} to illustrate the practical impact of Ca2KG on representative examples.

\subsection{Main Result (RQ1-RQ3)} 
\subsubsection{RQ1: Calibration and Utility} 
From Table~\ref{tab:main2}, we note that Ca2KG substantially improves calibration compared with baselines. Under both GPT-3.5 and LLaMA-3 backbones, Ca2KG consistently achieves the lowest ECE (e.g., 0.067/0.055 on MetaQA and 0.196/0.108 on WebQSP with GPT-3.5) and BS (0.078--0.128), while maintaining high AUC (up to 0.952). In contrast, KG-RAG and prompting baselines (Verb1S-Top4, Verb2S-Top4, Verb2S-CoT) exhibit poor calibration, with ECE values often exceeding 0.3--0.6. These results confirm that causality-aware approaches effectively reduce miscalibration while improving accuracy, demonstrating a superior balance of reliability and predictive performance.

\subsubsection{RQ2: Backbone Capacity} 
Comparing GPT-3.5 and LLaMA-3 results highlights the influence of the backbone LLM. GPT-3.5 generally achieves higher accuracies on WebQSP (e.g., 0.769 vs. 0.738 on 1-hop; 0.819 vs. 0.612 on 3-hop) and stronger calibration with lower ECE and BS, while LLaMA-3 performs competitively on MetaQA (e.g., 0.872 vs. 0.896 on 3-hop). These results suggest that stronger LLMs not only enhance reasoning accuracy but also improve calibration robustness when combined with Ca2KG. Importantly, the relative advantage of Ca2KG remains consistent across backbones, indicating that counterfactual prompting provides complementary benefits independent of backbone choice.

\subsubsection{RQ3: Task Difficulty}
The comparison between MetaQA and WebQSP illustrates the effect of task difficulty. MetaQA is synthetically generated and domain-specific, making it relatively easier, whereas WebQSP is based on real user queries, requiring semantic parsing and entity disambiguation, which substantially increases complexity. This difference is reflected in the results: even with GPT-3.5, accuracies on WebQSP are notably lower than on MetaQA (e.g., 0.769 vs. 0.876 in 1-hop, 0.819 vs. 0.896 in 3-hop), and calibration errors are consistently higher (ECE = 0.196/0.108 vs. 0.067/0.055). Nevertheless, Ca2KG maintains superior calibration and accuracy on both datasets, demonstrating robustness even in more challenging real-world QA settings.

\subsection{Efficiency Analysis (RQ4)}
\begin{figure}[t]
    \centering
    \begin{subfigure}{0.48\linewidth}
    \centering
    \includegraphics[width=\linewidth]{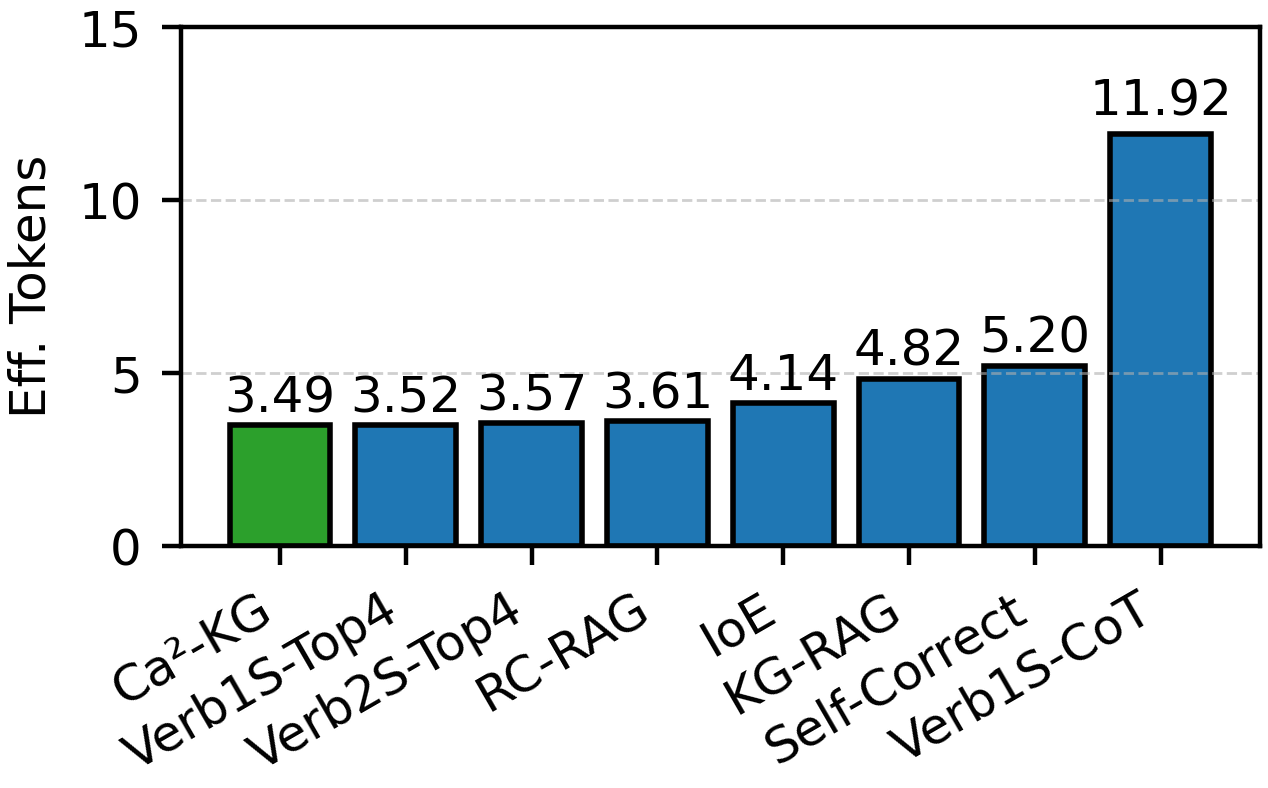}
    \subcaption{}
    \label{fig:eff1}
    \end{subfigure} 
    \hfill
    \begin{subfigure}{0.48\linewidth}
    \centering
    \includegraphics[width=\linewidth]{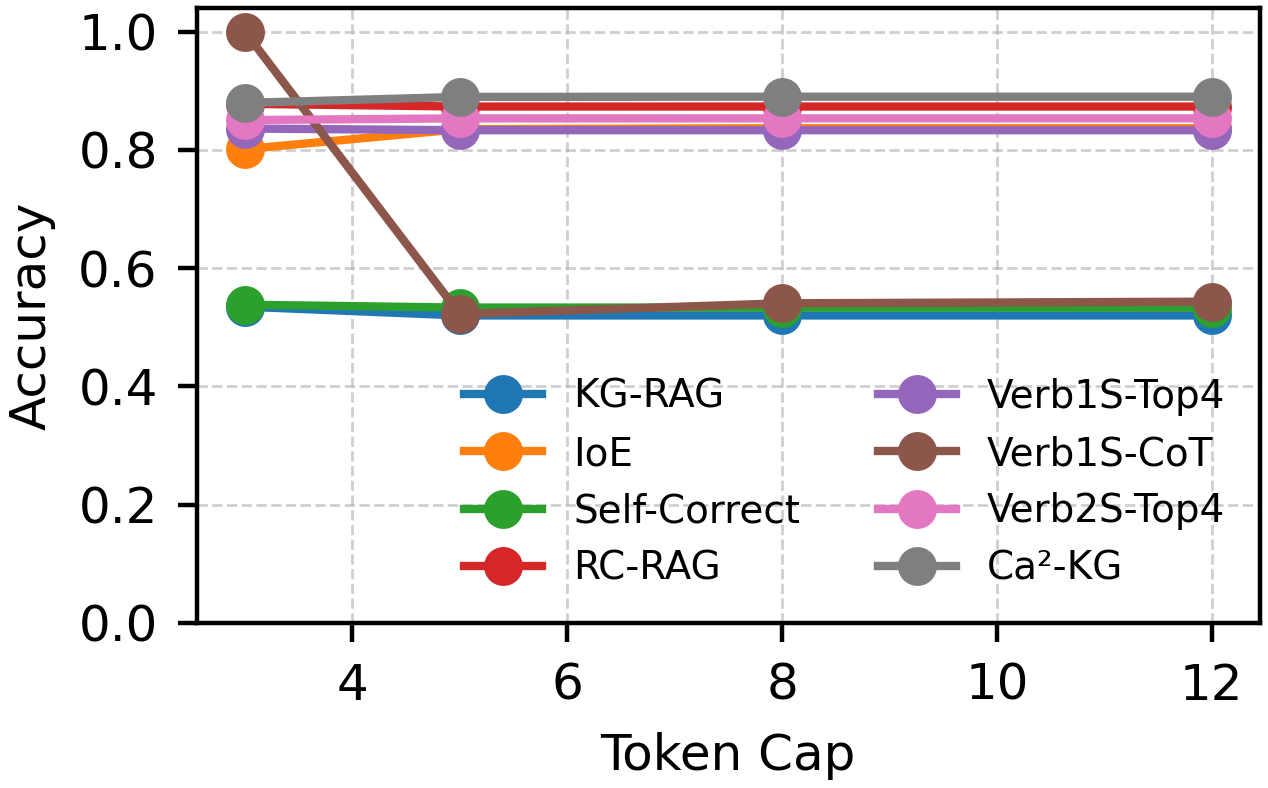}
    \subcaption{}
    \label{fig:eff2}
    \end{subfigure}
    \caption{Efficiency analysis on MetaQA. (a) Token usage per correct prediction in the 1-hop setting. (b) Accuracy under different token caps.}
    \label{fig:efficiency}
\end{figure}

We analyse the efficiency of different methods on MetaQA 1-hop in terms of token usage and performance under token caps. In Figure~\ref{fig:eff1}, Ca2KG achieves the lowest token cost per correct prediction (3.49), outperforming baselines such as KG-RAG (4.82), Self-Correct (5.20), and Verb2S-CoT (11.92). This shows that Ca2KG requires fewer tokens to achieve the same or better accuracy, making it highly cost-effective. Figure~\ref{fig:eff2} further confirms this advantage: when token budgets are restricted, Ca2KG maintains stable accuracy across different caps, in contrast to baselines such as KG-RAG (STD) and other prompting methods, which suffer sharp drops. These results demonstrate that counterfactual prompting not only improves calibration and accuracy (RQ1–RQ3), but also delivers strong efficiency, ensuring robustness under resource-constrained settings.

\begin{table*}[t]
\centering
\caption{\textbf{Ablation study on MetaQA and WebQSP.}}
\setlength{\tabcolsep}{3pt}
\begin{tabular}{l cccc cccc cccc cccc} 
\toprule
 & \multicolumn{8}{c}{\textbf{MetaQA}} & \multicolumn{8}{c}{\textbf{WebQSP}} \\
\cmidrule(lr){2-9}\cmidrule(lr){10-17}
 & \multicolumn{4}{c}{\textbf{1-hop}} 
 & \multicolumn{4}{c}{\textbf{3-hop}}
 & \multicolumn{4}{c}{\textbf{1-hop}}
 & \multicolumn{4}{c}{\textbf{3-hop}} \\
\cmidrule(lr){2-5}\cmidrule(lr){6-9}\cmidrule(lr){10-13}\cmidrule(lr){14-17}

\textbf{Method} 
& Acc$\uparrow$ & ECE$\downarrow$ & BS$\downarrow$ & AUC$\uparrow$ 
& Acc$\uparrow$ & ECE$\downarrow$ & BS$\downarrow$ & AUC$\uparrow$
& Acc$\uparrow$ & ECE$\downarrow$ & BS$\downarrow$ & AUC$\uparrow$
& Acc$\uparrow$ & ECE$\downarrow$ & BS$\downarrow$ & AUC$\uparrow$ \\
\midrule
Ca2KG
& 0.876 & 0.067 & 0.078 & 0.947 
& 0.896 & 0.055 & 0.069 & 0.950
& 0.769 & 0.196 & 0.184 & 0.837
& 0.819 & 0.108 & 0.128 & 0.920 \\

~w/o Path Quality $t_1$
& 0.866 & 0.103 & 0.080 & 0.946
& 0.881 & 0.075 & 0.080 & 0.936
& 0.768 & 0.207 & 0.217 & 0.826
& 0.808 & 0.202 & 0.135 & 0.823 \\

~w/o Rea. Reliability $t_2$
& 0.870 & 0.086 & 0.099 & 0.940
& 0.885 & 0.056 & 0.081 & 0.933
& 0.767 & 0.211 & 0.201 & 0.808
& 0.803 & 0.209 & 0.143 & 0.837 \\

~w/o Both ($t_1$ \& $t_2$)
& 0.865 & 0.134 & 0.134 & 0.894
& 0.817 & 0.172 & 0.172 & 0.825
& 0.752 & 0.245 & 0.245 & 0.753
& 0.791 & 0.213 & 0.175 & 0.816 \\
\bottomrule
\end{tabular}
\label{tab:ablation}
\end{table*}

\subsection{Ablation Study (RQ5)}
Table~\ref{tab:ablation} shows that both Path Quality Intervention ($t_{1}$) and Reasoning Reliability Intervention ($t_{2}$) play complementary roles in enhancing model effectiveness. Removing $t_{1}$ leads to clear degradation in calibration, with ECE and BS increasing across datasets (e.g., ECE rises from 0.067 to 0.103 on MetaQA 1-hop, and from 0.108 to 0.202 on WebQSP 3-hop), and AUC dropping (0.950 $\rightarrow$ 0.936). Similarly, excluding $t_{2}$ causes a moderate decline in performance, particularly in calibration metrics (e.g., BS = 0.099 vs. 0.078 on MetaQA 1-hop; ECE = 0.211 vs. 0.196 on WebQSP 1-hop). Notably, removing both interventions results in the most severe degradation: accuracy decreases substantially (e.g., 0.896 $\rightarrow$ 0.817 on MetaQA 3-hop, 0.769 $\rightarrow$ 0.752 on WebQSP 1-hop), calibration errors nearly double (ECE = 0.067 $\rightarrow$ 0.134 on MetaQA 1-hop, 0.108 $\rightarrow$ 0.213 on WebQSP 3-hop), and AUC declines sharply (0.950 $\rightarrow$ 0.825). These results demonstrate that $t_{1}$ is especially crucial for calibration and robustness, $t_{2}$ further improves reliability and uncertainty estimation, and together they provide complementary benefits that are critical for achieving state-of-the-art performance.

\section{Related Work}
\subsection{Calibration of LLM}
Current strategies for improving calibration in LLMs span a wide spectrum. Several studies~\cite{lin2022teaching,park2022calibration,kuhn2023semantic,xiao2022uncertainty} demonstrate that combining large pre-trained models with temperature scaling~\cite{guo2017calibration} can yield better calibrated predictions. Other works explore whether linguistic cues in model outputs provide reliable signals of uncertainty, and how these align with actual confidence~\cite{zhou2023navigating,mielke2022reducing,tian2023just}. More recently, prompting-based approaches have attracted attention due to their flexibility and applicability to black-box LLMs, enabling techniques such as self-reported confidence, multi-sample self-consistency, and explicit uncertainty expression. These are especially valuable for open-ended or generative tasks where post-hoc calibration is less effective~\cite{shorinwa2025survey}. For KG-RAG, prompting-based strategies are particularly appealing as they allow models to reason over structured knowledge paths and assess evidence consistency without modifying model architectures. Distinct from prior work, we introduce counterfactual prompting to explicitly simulate failures in knowledge retrieval and reasoning.

\subsection{KG-RAG}
Recent efforts have integrated knowledge graphs into retrieval-augmented generation to improve factual accuracy, reasoning, and interpretability. For example, \citet{hu2025grag} propose GRAG, which retrieves relevant sub-graphs to guide multi-hop QA; \citet{liang2025kag} introduce KAG, which enhances alignment between triples and text with an attention-guided encoder-decoder; \citet{liu2024knowledge} investigate knowledge-injected prompting strategies that verbalise KG paths into natural language; and \citet{tan2025paths} present a path-level reasoning framework that explicitly models retrieval confidence and path reliability. These approaches have substantially advanced KG-RAG by improving the grounding of LLM outputs, enabling more precise entity disambiguation and more controllable reasoning over structured knowledge. However, most of them concentrate on improving factual accuracy and reasoning capability, while paying limited attention to the equally critical issue of calibration. We argue that carefully designed prompting strategies can guide KG-RAG models to introspect on conflicting or missing evidence, and in doing so, provide confidence estimates that are more calibrated, interpretable, and ultimately more reliable for real-world use.


\subsection{Causal Inference for LLM}
Causal inference aims to uncover the mechanisms underlying variable interactions through rigorous methodologies~\cite{pearl2016causal}. Building on solid theoretical foundations, many approaches have been developed to estimate causal effects~\cite{du2025telling,XuCLLLW23,ZhangXCLLXF25}, even in the presence of unobserved confounders~\cite{10791303,DBLP:conf/iclr/XuCLL0Y24,hengXL0LL24}. These techniques have been widely applied in NLP, including de-biasing~\cite{zhao2025unbiased}, fake news detection~\cite{wang2022causal}, and sentiment analysis~\cite{abs-2507-00389}. More recently, researchers have begun incorporating causal reasoning into prompting. For example, Causal Prompting~\cite{zhang2025causal} formulates prompts based on hypothesised causal structures to elicit causally consistent outputs, while DeCoT~\cite{DBLP:conf/acl/Wu0CWRKRM24} embeds causal structures into chain-of-thought reasoning, using front-door adjustment and instrumental variables to mitigate spurious reasoning caused by hidden confounders. Inspired by these advances, our proposed Ca2KG framework introduces a new perspective by integrating causal reasoning with knowledge graph–based retrieval. In particular, Ca2KG leverages structured causal paths from knowledge graphs to construct counterfactually informed prompts, enabling LLMs to not only generate answers but also provide uncertainty-aware and causally grounded reasoning.

\section{Conclusion}
In this paper, we present {Ca2KG}, a causality-aware calibration framework for Knowledge Graph Retrieval-Augmented Generation. By interpreting counterfactual prompting strategies as causal interventions and combining them with a panel-based re-scoring mechanism, our framework exposes and quantifies retrieval-dependent uncertainties in both knowledge quality and reasoning reliability. Extensive experiments on MetaQA and WebQSP show that Ca2KG consistently reduces overconfidence and achieves state-of-the-art calibration performance, while maintaining or even enhancing predictive accuracy. Beyond addressing the calibration gap in KG-RAG, our work highlights the importance of causality-inspired prompting as a general strategy for improving trustworthiness in web-scale knowledge retrieval and generation.

\bibliographystyle{ACM-Reference-Format}
\balance
\bibliography{sample-base}

\appendix

\section{Ethical Use of Data and Informed Consent}

This work uses two widely adopted benchmark datasets, MetaQA and WebQSP, which are publicly available and have been extensively used in prior research on knowledge graph question answering. Both datasets are released under open licenses for research purposes and do not contain personally identifiable information or sensitive data. No new data involving human participants were collected in this work, and therefore no Institutional Review Board (IRB) approval was required. All experiments are conducted in accordance with ACM’s Publications Policy on Research Involving Human Participants and Subjects.
	
\section{Case Study}
\label{Case}

\promptbox{Case Study I on WebQSP}{
\textbf{Question:}~what is the capital of argentina?\\
\textbf{Ground-truth:}~Buenos Aires \\

\textbf{Step 1: Counterfactual Prompting}

\begin{itemize}[leftmargin=*,noitemsep]
  \item \textbf{initial prompt $t_{0}$}: \\
  Use the provided contexts to answer the question. If the contexts are incomplete or weak, still provide your best possible answer. Output MUST be exactly one line in this format: \{final answer\}. Do not include any other text. Examples: \{Italian\} \\
  \textbf{Answer $a_{0}$}: \texttt{argentina}

  \item \textbf{Path Quality Intervention $t_{1}$}: \\
  Assume your previous answer is wrong because the quality of the referred contexts is poor. Re-select the most relevant parts from the given contexts and regenerate the answer using one or a few words. Output MUST be exactly one line in this format: \{final answer\}. Do not include any other text. Examples: \{Italian Languages\} \\
  \textbf{Answer $a_{1}$}: \texttt{buenos aires}

  \item \textbf{Reasoning Reliability Intervention $t_{2}$}: \\
  Assume your previous answer is wrong due to improper use of the retrieved contexts. Carefully re-check the provided contexts and regenerate the answer using one or a few words. Output MUST be exactly one line in this format:\{final answer\}. Do not include any other text. Examples: \{Italian Languages\} \\
  \textbf{Answer $a_{2}$}: \texttt{buenos aires}
\end{itemize}

\medskip
\textbf{Step 2: Panel-based Re-scoring}

You are given 3 candidate answers for the same question from different prompts: $t_{0}$: $a_{0}$; $t_{1}$: $a_{1}$; $t_{2}$: $a_{2}$. \\ 
Task: (1) Build a unified set of UNIQUE answers by merging semantically identical strings across $a_{0}$, $a_{1}$, $a_{2}$. (2) Compute a GLOBAL frequency vector over the unique answers based on how many of $a_{0}$/$a_{1}$/$a_{2}$ map to each canonical answer. (3) Assign probabilities for each scenario ($t_{0}$, $t_{1}$, $t_{2}$) to be exactly this global frequency vector normalised by 3 (counts/3), after duplicate aggregation and before rounding. Merging rules: ignore case, whitespace, punctuation, and plural/singular differences; choose a clean canonical form for answers; if multiple inputs map to the same canonical answer, aggregate by summing before normalisation. Probability rules: A) Let count[i] be how many of \{$a_{0}$, $a_{1}$, $a_{2}$\} map to answers[i]. Then for every scenario S in \{$t_{0}$, $t_{1}$, $t_{2}$\}, set S[i] = round(count[i]/3, 2). B) Do not output equal probabilities across answers when counts differ. C) Each probability $\in [0.00, 1.00]$; sums may be $< 1.00$ after rounding. D) If a scenario has no plausible answers, use zeros. Return exactly one line of strict JSON, no wrapper, no extra text.\\

\textbf{Canonical merge:} \texttt{Buenos Aires}, \texttt{Argentina}\\
\textbf{Counts:} \texttt{Buenos Aires = 2}, \texttt{Argentina = 1}\\
\textbf{Verbalised Prior:} \\\texttt{"Buenos Aires": 0.62, "Argentina": 0.38}

\medskip
\textbf{Step 3: Accurate Answer Generation}

\textbf{Panel JSON: }\\
\texttt{\detokenize{{"answers":["buenos aires","argentina"],}}} \\
\texttt{\detokenize{{"t_0":[0.62,0.38],}}}\\
\texttt{\detokenize{{"t_1":[0.68,0.32],}}}\\
\texttt{\detokenize{{"t_2":[0.80,0.20]}}}

\textbf{Causal effect variation ($\text{CE}_{\text{var}}$):}
\begin{align*}
\text{CE}_{\text{var}}&(\text{Buenos Aires}) \\
  &= \max(0.62, 0.68, 0.80) - \min(0.62, 0.68, 0.80) \\
  &= 0.80 - 0.62 = 0.18
\end{align*}
\begin{align*}
\text{CE}_{\text{var}}&(\text{Argentina}) \\
  &= \max(0.38, 0.32, 0.20) - \min(0.38, 0.32, 0.20) \\
  &= 0.38 - 0.20 = 0.18
\end{align*}

\textbf{Average causal effect ($\overline{\text{CE}}$):}
\begin{align*}
\overline{\text{CE}}(\text{Buenos Aires}) = (0.62 + 0.68 + 0.80)/3 = 0.70
\end{align*}
\begin{align*}
\overline{\text{CE}}(\text{Argentina}) = (0.38 + 0.32 + 0.20)/3 = 0.30
\end{align*}

\textbf{Causal calibration index ($\text{CCI}$):}
\begin{align*}
\text{CCI}(\text{Buenos Aires}) = 0.70 \times (1 - 0.18) = 0.70 \times 0.82 = 0.57
\end{align*}
\begin{align*}
\text{CCI}(\text{Argentina}) = 0.30 \times (1 - 0.18) = 0.30 \times 0.82 = 0.25
\end{align*}

\textbf{Final Answer:~}\texttt{\{Buenos Aires\}}
}

\promptbox{Case Study II on MetaQA}{
\textbf{Question:}~The films that share directors with the films [Tiresia] are written by who?\\
\textbf{Ground-truth:}~Bertrand Bonello \\

\textbf{Step 1: Counterfactual Prompting}

\begin{itemize}[leftmargin=*,noitemsep]
  \item \textbf{initial prompt $t_{0}$}: \\
  Use the provided contexts to answer the question. If the contexts are incomplete or weak, still provide your best possible answer. Output MUST be exactly one line in this format: \{final answer\}. Do not include any other text. Examples: \{Italian\} \\
  \textbf{Answer $a_{0}$}: \texttt{Luca Fazzi}

  \item \textbf{Path Quality Intervention $t_{1}$}: \\
  Assume your previous answer is wrong because the quality of the referred contexts is poor. Re-select the most relevant parts from the given contexts and regenerate the answer using one or a few words. Output MUST be exactly one line in this format: \{final answer\}. Do not include any other text. Examples: \{Italian Languages\} \\
  \textbf{Answer $a_{1}$}: \texttt{Bertrand Bonello}

  \item \textbf{Reasoning Reliability Intervention $t_{2}$}: \\
  Assume your previous answer is wrong due to improper use of the retrieved contexts. Carefully re-check the provided contexts and regenerate the answer using one or a few words. Output MUST be exactly one line in this format:\{final answer\}. Do not include any other text. Examples: \{Italian Languages\} \\
  \textbf{Answer $a_{2}$}: \texttt{Bertrand Bonello}
\end{itemize}

\medskip
\textbf{Step 2: Panel-based Re-scoring}

You are given 3 candidate answers for the same question from different prompts: $t_{0}$: $a_{0}$; $t_{1}$: $a_{1}$; $t_{2}$: $a_{2}$. \\ 
Task: (1) Build a unified set of UNIQUE answers by merging semantically identical strings across $a_{0}$, $a_{1}$, $a_{2}$. (2) Compute a GLOBAL frequency vector over the unique answers based on how many of $a_{0}$/$a_{1}$/$a_{2}$ map to each canonical answer. (3) Assign probabilities for each scenario ($t_{0}$, $t_{1}$, $t_{2}$) to be exactly this global frequency vector normalised by 3 (counts/3), after duplicate aggregation and before rounding. Merging rules: ignore case, whitespace, punctuation, and plural/singular differences; choose a clean canonical form for answers; if multiple inputs map to the same canonical answer, aggregate by summing before normalisation. Probability rules: A) Let count[i] be how many of \{$a_{0}$, $a_{1}$, $a_{2}$\} map to answers[i]. Then for every scenario S in \{$t_{0}$, $t_{1}$, $t_{2}$\}, set S[i] = round(count[i]/3, 2). B) Do not output equal probabilities across answers when counts differ. C) Each probability $\in [0.00, 1.00]$; sums may be $< 1.00$ after rounding. D) If a scenario has no plausible answers, use zeros. Return exactly one line of strict JSON, no wrapper, no extra text.\\

\textbf{Canonical merge:} \texttt{Bertrand Bonello}, \texttt{Luca Fazzi}\\
\textbf{Counts:} \texttt{Bertrand Bonello = 2}, \texttt{Luca Fazzi = 1}\\
\textbf{Verbalised Prior:} \\\texttt{"Bertrand Bonello": 0.67, "Luca Fazzi": 0.33}

\medskip
\textbf{Step 3: Accurate Answer Generation}

\textbf{Panel JSON: }\\
\texttt{\detokenize{{"answers":["luca fazzi","bertrand bonello"],}}} \\
\texttt{\detokenize{{"t_0":[0.33,0.67],}}}\\
\texttt{\detokenize{{"t_1":[0.33,0.67],}}}\\
\texttt{\detokenize{{"t_2":[0.33,0.67]}}}

\textbf{Causal effect variation ($\text{CE}_{\text{var}}$):}
\begin{align*}
\text{CE}_{\text{var}}&(\text{Bertrand Bonello}) \\
  &= \max(0.67, 0.67, 0.67) - \min(0.67, 0.67, 0.67) \\
  &= 0.00
\end{align*}
\begin{align*}
\text{CE}_{\text{var}}&(\text{Luca Fazzi}) \\
  &= \max(0.33, 0.33, 0.33) - \min(0.33, 0.33, 0.33) \\
  &= 0.00
\end{align*}

\textbf{Average causal effect ($\overline{\text{CE}}$):}
\begin{align*}
\overline{\text{CE}}(\text{Bertrand Bonello}) 
  &= (0.67 + 0.67 + 0.67)/3 = 0.67
\end{align*}
\begin{align*}
\overline{\text{CE}}(\text{Luca Fazzi}) 
  &= (0.33 + 0.33 + 0.33)/3 = 0.33
\end{align*}

\textbf{Causal calibration index ($\text{CCI}$):}
\begin{align*}
\text{CCI}(\text{Bertrand Bonello}) 
  &= 0.67 \times (1 - 0.00) = 0.67
\end{align*}
\begin{align*}
\text{CCI}(\text{Luca Fazzi}) 
  &= 0.33 \times (1 - 0.00) = 0.33
\end{align*}

\textbf{Final Answer:~}\texttt{\{Bertrand Bonello\}}
}

\end{document}